\documentclass[journal]{IEEEtran}
\usepackage{amsmath, amssymb, amsthm}
\usepackage{hyperref}
\usepackage{cite}
\usepackage{graphicx}
\usepackage{xfrac}
\usepackage{stfloats}
\usepackage{tikz}
\usepackage{multirow}
\usepackage{tabularx,booktabs}
\usepackage{siunitx}
\usepackage{algorithm}
\usepackage{algpseudocode}
\newcolumntype{C}{>{\centering\arraybackslash}X} 
\newcolumntype{R}{>{\raggedleft\arraybackslash}X} 
\newcolumntype{L}{>{\raggedright\arraybackslash}X} 
\newcolumntype{P}[1]{>{\centering\arraybackslash}p{#1}}

\usepackage{pgfplots}
\pgfplotsset{compat=newest}
\usepackage{svg}
\usepackage[dvipsnames]{xcolor}
\newtheorem*{objective*}{Objective}

\newtheorem{example}{Example}
\newtheorem{remark}{Remark}

\DeclareSymbolFont{cmbrmath}{OML}{cmbrm}{m}{it}
\DeclareSymbolFont{cmbrmathbf}{OML}{cmbrm}{b}{it}
\DeclareSymbolFont{cmbrupright}{OT1}{cmbr}{b}{n}

\DeclareMathSymbol{\cmbtheta}{\mathord}{cmbrmath}{"12}
\DeclareMathSymbol{\cmbphi}{\mathord}{cmbrmath}{"1E}
\DeclareMathSymbol{\cmbrho}{\mathord}{cmbrmath}{"1A}
\DeclareMathSymbol{\cmbv}{\mathalpha}{cmbrmath}{`v}
\DeclareMathSymbol{\cmbbz}{\mathalpha}{cmbrupright}{`z}
\DeclareMathSymbol{\cmbby}{\mathalpha}{cmbrupright}{`y}

\definecolor{myblue}{rgb}{0.06,0.258,0.501}
\definecolor{myred}{rgb}{1, 0.27, 0.227}
\definecolor{mygreen}{RGB}{34,144,82}

\begin{document}

\title{Risk Estimation for Automated Driving}

\author{Leon Tolksdorf$^{1,2}$, Arturo Tejada$^{1, 3}$, Jonas Bauernfeind$^{2}$, Christian Birkner$^{2}$, and Nathan van de Wouw$^{1}$
\thanks{$^{1}$Department of Dynamics and Control, Eindhoven University of Technology, Eindhoven, The Netherlands, e-mail:
        {\tt\small \{l.t.tolksdorf, a.tejada.ruiz, n.v.d.wouw\}@tue.nl}}%
\thanks{$^{2}$CARISSMA Institute of Safety in Future Mobility, Technische Hochschule Ingolstadt, Ingolstadt, Germany, e-mail:
        {\tt\small \{leon.tolksdorf, jonas.bauernfeind, christian.birkner\}@thi.de}}%
\thanks{$^{3}$TNO, Integrated Vehicle Safety, Helmond, The Netherlands, e-mail:
        {\tt\small arturo.tejadaruiz@tno.nl}}%
}

\maketitle

\begin{abstract}
Safety is a central requirement for automated vehicles. As such, the assessment of risk in automated driving is key in supporting both motion planning technologies and safety evaluation. In automated driving, risk is characterized by two aspects. The first aspect is the uncertainty on the state estimates of other road participants by an automated vehicle. The second aspect is the severity of a collision event with said traffic participants. Here, the uncertainty aspect typically causes the risk to be non-zero for near-collision events. This makes risk particularly useful for automated vehicle motion planning. Namely, constraining or minimizing risk naturally navigates the automated vehicle around traffic participants while keeping a safety distance based on the level of uncertainty and the potential severity of the impending collision. Existing approaches to calculate the risk either resort to empirical modeling or severe approximations, and, hence, lack generalizability and accuracy. In this paper, we combine recent advances in collision probability estimation with the concept of collision severity to develop a general method for accurate risk estimation. The proposed method allows us to assign individual severity functions for different collision constellations, such as, e.g., frontal or side collisions. Furthermore, we show that the proposed approach is computationally efficient, which is beneficial, e.g., in real-time motion planning applications. The programming code for an exemplary implementation of Gaussian uncertainties is also provided.
\end{abstract}


\IEEEpeerreviewmaketitle

\section{Introduction}
Driving safely towards a goal destination is a key functionality in automated vehicle (AV) design. However, assessing safety is an intricate matter, as even a motion plan that is collision free may not feel safe for the passengers of the AV. Furthermore, determining whether a (future) motion plan is collision free requires precise knowledge of the current and future states of obstacles and other road participants. Such knowledge, however, is generally only partial in nature as measurement and estimation introduces uncertainties. To account for uncertainties, risk functions are commonly used to assess the safety of a motion plan, where risk is understood as stochastic quantity describing potential harm given limited knowledge, i.e., uncertainty, about current and future driving states. \\
Risk functions based on this \textit{rough} definition are typically constituted of two aspects: one related to the uncertainty on current and future driving states and the other to the severity of an interaction, e.g., a collision. Such risk measures have been shown to correlate with perceived risk by humans \cite{kolekar2020human, he2024new, krugel2025international}, where \cite{kolekar2020human} has also shown that constraining risk within a motion planning algorithm in a static environment (i.e., without other dynamic actors) led to human-comparable vehicle behavior. Furthermore, minimizing or constraining risk within a motion planner is established to generate safe AV behavior (see, e.g., \cite{mustafa2024racp,tolksdorf2023risk, nyberg2021risk, schwarting2017safe, ploeg2024risk}).\\
A key challenge is \textit{computationally efficient} risk estimation, which is generally desired in real-time motion planning applications. As analytic methods are scarce, some works (see, e.g., \cite{ploeg2024risk, kolekar2020parts, he2024new, wang2016driving, dixit2019trajectory}) resort to modelling risk by empirically parameterizing model equations, thus not guaranteeing that the uncertainty is correctly characterized in a probabilistic sense, and generalizability beyond the scope of data used for tuning the model is unclear. Other works choose to over-approximate the risk by representing each actor by a circular shape (see, e.g., \cite{mustafa2024racp, tolksdorf2023risk}), which simplifies the calculations significantly; however, it clearly undermines the accuracy of estimation. Furthermore, in \cite{tolksdorf2023risk, schreier2016integrated, Tolksdorf_SAFE_UP_2024}, the risk is calculated with Monte Carlo sampling, which itself is causing a significant computational burden. An important observation is that if the severity term is constant, one is left to compute the probability of collision. Recently, \cite{Tolksdorf_POC_2024} proposes a new method based on a multi-circular shape approximation of each actor to estimate the collision probability. That method is particularly tailored to optimization-based motion planning as it is guaranteed to over-approximate the collision probability value while being computationally efficient and supporting the solver in generating smooth, reproducible trajectories. From a risk estimation perspective, modeling a vehicle by individual areas, e.g., multiple overlapping circles, is a promising approach not only for uncertainty estimation but also for the calculation of the severity, since the collision severity is known to depend on the location of impact for each actor, i.e., the collision constellation \cite{abdel2005exploring, D2.6, D4.1}.\\
This paper builds upon the multi-circular approach for collision probability estimation of \cite{Tolksdorf_POC_2024} to develop a novel and accurate methodology for risk estimation. This approach, in which all road actors are represented by multiple overlapping circles, allows to assign different severity functions for each pair of colliding circles. Here, a particular circle may be related to the front, middle, rear, etc., of a vehicle, and, hence, e.g., a front-to-side collision can be represent by assigning individual severity functions for such a circle-to-circle pair. To combine many individual severity functions, we introduce an averaging method to combine individual severity functions to accurately represent different collision constellations by a single severity value. Our approach retains the over-approximate guaranties in the uncertainty aspect and also inherits the advantageous properties for optimization-based motion planning from~\cite{Tolksdorf_POC_2024}, by using the same algorithmic structure and multi-circular approach.
Finally, we present an exemplary implementation of the proposed risk estimation approach for Gaussian uncertainties and a kinetic energy model to assign collision severity. The resulting algorithm is computing sufficiently fast to support real-time motion planning applications and is provided in open-source.\\
The remainder of this article is structured as follows. Section \ref{sec:problem_statement} introduces preliminary notation and states the problem of risk estimation. Following, Section \ref{sec:methodlogy} derives the risk based on a multi-cirucular shape approximation that over-approximates the uncertainty aspects and includes a model to combine individual severity functions to account for different collision constellations. Section \ref{sec:risk_for_gaussian_uncertainties} presents an implementation for Gaussian uncertainties with a kinematic-energy severity model, for which we then demonstrate the risk estimation in Section \ref{sec:numerical_case_studies}. Lastly, we conclude this article in Section \ref{sec:conclusion}.

\section{Problem Statement}\label{sec:problem_statement}
\begin{figure}[t!]
\begin{center}
\includegraphics[width=8.4cm]{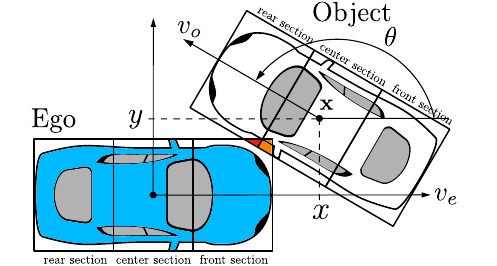}  
\caption{Problem Statement: The ego (blue vehicle) is colliding with its front section into the object's (white vehicle) front and side section colored red and orange, respectively.}
\label{fig:problem_statement}
\end{center}
\end{figure}
Consider an arbitrary traffic scene with two actors: an automated vehicle, called the ego,  and another road user (e.g., another vehicle or a cyclist), called the object, see Figure~\ref{fig:problem_statement}. For clarity of exposition, we restrict ourselves to only one object; however, our method also applies to any number of objects. From the perspective of the ego, the object is characterized by its configuration $\mathbf{y} := (\mathbf{x}, \theta)\in
\mathcal{C} := \mathbb{R}^3$, which is composed of the Cartesian position of the object's geometrical center with respect to the ego's geometrical center, $\mathbf{x} := (x, y) \in\mathbb{R}^2$, and the object's heading angle measured with respect to the ego's longitudinal axis, $\theta \in \mathbb{R}$ ($\mathcal{C}$ is called the configuration space). Note that we define $\theta$ on $\mathbb{R}$, as for later calculations it is beneficial to represent angles as multiples of $2\pi$. Figure \ref{fig:problem_statement} depicts all the described coordinates. In the sequel, a configuration at the discrete-time instant $k$ is denoted by $\mathbf{y}_{k}$.\\
We define risk as the expected severity of a risk-inducing event $\mathcal{E}$ (e.g., collisions or violating traffic regulation), as in~\cite{tolksdorf2023risk}. In this paper, we only consider collisions and, hence, our objective is to estimate collision risk. To estimate the expected severity of a collision (i.e., collision risk), one needs to derive the conditions on the relative position and heading angle of the vehicles that, when satisfied, imply the vehicles are colliding and, if so, what the severity of that collision is.    \\
Let $\mathbf{z}_k \in \mathbb{R}^{n_\mathbf{z}}$ denote the variables involved in determining the occurrence and severity of a collision event $\mathcal{E}$ at time $k$, these are, e.g., positions, orientations, and velocities of both actors. In general $\mathbf{z}_k$ is a vector of random variables with associated probability density function (PDF) $p_{\mathbf{z}}(\cdot ;\mathbf{w}_k)$. Here, we assume that the random variables are independent and that $p_{\mathbf{z}}$ has the same functional form for all $k$, but time-varying parameters denoted by $\mathbf{w}_k$ (e.g., time varying mean and variances), which characterizes the time dependence of the PDF. 
Let the event set $\mathcal{B}_{\mathcal{E}} \subset \mathbb{R}^{n_\mathbf{z}}$ be such that if $\mathbf{z}_k\in \mathcal{B}_{\mathcal{E}}$, then the event $\mathcal{E}$, i.e., a collision, has occurred. Note that in order to determine whether a collision has occurred, only information about the configuration $\mathbf{y}$ is required, as $\mathbf{y}$ suffices to check if both actors' footprints intersect. However, to estimate the severity, additional information about the kinematic variables of both actors, e.g., the velocities $v_e, v_o$, see Figure~\ref{fig:problem_statement}, may also be needed.\\
Based on accidentology research, see e.g., \cite{abdel2005exploring, D2.6, D4.1}, we assume that the severity of a collision depends on the collision constellation, i.e., the specific sections of each vehicle which are colliding. Therefore, consider that the ego vehicle consists of $N_e$ sections (e.g., $N_e = 3$: front, center, rear) and the object is constituted of $N_o$ sections. To each specific section pair (e.g., ego front section to object rear section), we allocate an individual severity function that assigns the severity for that section pair, based on the actors' velocities $v_e, v_o$ and their relative directionality $\theta$, see the collisions among different vehicular sections in Figure \ref{fig:problem_statement}. Hence, to the $j$-th ego section, $j\in \{1,2, ...N_e\}$, colliding with the $l$-th object section, $l\in \{1,2,...,N_o\}$, we assign the severity with $s_{j,l}$ such that $(\theta, v_e, v_o) \mapsto s_{j,l}(\theta, v_e, v_o) \in \mathbb{R}_{\geq0}$. Moreover, we collect all individual severity functions in a set $\mathcal{S} = \{ s_{j,l} \mid \forall j \in \{1, 2, \dots, N_e\}, \forall l\in \{1, 2, \dots, N_o\}\}$. With the collection of all severity functions in $\mathcal{S}$, we define a function $f_s$ that combines all $N_eN_o$ individual severity functions into a total severity function. In practice, this function may, e.g., represent the sum of all individual severities or the maximum severity. To retain generality, we define the total severity as 
\begin{equation}\label{eq:total_severity}
    s = f_s(\mathcal{S}), \text{ such that } s \geq 0.
\end{equation}
As a result, an element $\mathbf{z} \in \mathcal{B}_{\mathcal{E}}$ contains $x, y, \theta, v_e, v_o$, of which $x, y, \theta$ are needed for determination of collision probability and $\theta, v_e, v_o$ are used to assess the severity. Note that although $\theta$ is needed for both, its range for which the event $\mathcal{E}$ occurs is restricted by the set $\mathcal{A}_{rec} \in \mathcal{C}$ of configurations that lead to a collision among two vehicles' rectangular footprints. Finally, we assume that all kinematic variables associated with the ego vehicle are known without any uncertainty, thus, $v_e$ will be treated as a parameter to our risk estimation.
Hence, $\mathcal{B}_\mathcal{E} := \mathcal{A}_{rec} \times \mathcal{V}_o$, where $\mathcal{V}_o$ is a subset of $\mathbb{R}$ and represents the possible collision velocities of the object.
Therefore, the risk $R_k$ associated with $\mathcal{E}$ at time $k$ is defined as 
\begin{equation}\label{eq:risk_general_1}
    R^{rec}(v_e,\mathbf{w}_k) :=  \int_{\mathcal{B}_{\mathcal{E}}}s(\cmbtheta, v_e, \cmbv_{o})p_{\mathbf{z}}(\cmbbz;\mathbf{w}_k)\mathrm{d}\cmbbz.
\end{equation}
Note that we distinguish integration variables with the computer modern bright font, hence, we use $\cmbtheta, \cmbv_{o}, \cmbbz$ instead of $\theta,v_o, \mathbf{z}$ in (\ref{eq:risk_general}), as we integrate over all random variables in $\mathbf{z}$.
The objective of this paper is to estimate $R_k$ in (\ref{eq:risk_general_1}) by over-approximating both actors' shapes with multiple overlapping circles, giving a larger set of possible collision configurations, i.e., $\mathcal{A}_{cir} \supset \mathcal{A}_{rec}$. The over-approximation is necessary, as the multi-circular approach allows for a more efficient computation than rectangular approaches, and under-approximation is generally undesired in safety-critical applications. Here, the multi-circular over-approximation naturally yields the individual severity functions $s_{j,l}$ by considering individual ego and object circle pairs, e.g., a front object circle colliding with a rear ego circle may represent a rear end collision. The individual severity functions $s_{j,l}$ are then combined by $f_s$ in (\ref{eq:total_severity}). Thus, the domain of integration for the multi-circular shape representation is given by  $\tilde{\mathcal{B}}_\mathcal{E} := \mathcal{A}_{cir} \times \mathcal{V}_o$. Hence, our objective is to estimate
\begin{equation}\label{eq:risk_general}
        R^{cir}(v_e; \mathbf{w}_k) :=  \int_{\tilde{\mathcal{B}}_{\mathcal{E}}}s(\cmbtheta, v_e, \cmbv_{o})p_{\mathbf{z}}(\cmbbz;\mathbf{w}_k)\mathrm{d}{\cmbbz}.
\end{equation}

\section{Methodology for Risk Estimation}\label{sec:methodlogy}
In order to determine the risk at a certain time instant $k$, the integral in (\ref{eq:risk_general}) needs to be evaluated. In general, analytic solutions are not available, hence, the risk is either over-approximated \cite{schwarting2017safe} or estimated with random sampling, i.e., Monte Carlo methods \cite{schreier2016integrated, tolksdorf2023risk}. In the sequel, we will construct the risk for an averaged sum of severities in (\ref{eq:total_severity}), which is averaged in cases a collision among multiple circles occurs. Since the risk is estimated in the same way for all times $k$, we omit time indexing in the sequel for notational clarity. As we assume independence of the random variables in $\mathbf{z}$, we can decompose the PDF into its components as $p_{\mathbf{z}}(\mathbf{z};\mathbf{w}) = p_{\mathbf{y}}(\mathbf{y}; \mathbf{w}_{\mathbf{y}})p_{v_o}(v_o;\mathbf{w}_{v_o}) = p_{\mathbf{x}}(x, y, \mathbf{w}_{\mathbf{x}})p_{\theta}(\mathtt{\theta}; \mathbf{w}_{\theta})p_{v_o}(v_o;\mathbf{w}_{v_o})$. 
\subsection{From Collision Probability to Risk of Collision}
In fact, computing the risk in (\ref{eq:risk_general}) is closely related to calculating the probability of collision between a vehicle and an object. Considering the relative object configuration $\mathbf{y}$ to be uncertain and $\mathcal{A}_{rec}$ to be the set of all relative object configurations leading to a collision with the ego, the probability of collision is given by  
\begin{equation}\label{eq:general_POC}
    \mathbb{P}\{\mathbf{y} \in\mathcal{A}_{rec}\} = \int\displaylimits_{ \mathcal{A}_{rec}} p_{\mathbf{y}}(\cmbby; \mathbf{w}_{\mathbf{y}}) \text{d} \cmbby.
\end{equation}
In \cite{Tolksdorf_POC_2024}, (\ref{eq:general_POC}) is over-approximated by enclosing both the vehicle's and object's rectangular footprints in multiple overlapping circles. This approach allows for fast computation of the collision probability, while guaranteeing over-approximation. We will extend that approach from collision probability estimation to risk estimation.
\begin{figure}[t!]
\begin{center}
\includegraphics[width=8.4cm]{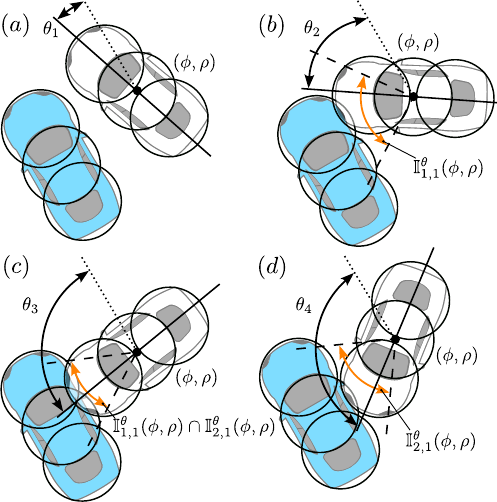}  
\caption{Example: Intersection angle intervals. In all subfigures, the object (white vehicle) is located at the same relative position $(\phi, \rho)$ to the ego (blue vehicle).}
\label{fig:severity_intervals}
\end{center}
\end{figure}
\\Suppose that the vehicle and object shapes are (tightly) covered by $N_{e}$ and $N_{o}$ overlapping circles with radii $r_e, r_o$, respectively (see Figure \ref{fig:severity_intervals} (a)). The circles of each actor are placed equidistantly apart by distances $d_{e}$ and $d_{o}$ between the circle centers for the ego and object, respectively. Then, the furthest distance between both actors' geometric centers, for which a collision of the multi-circular shape approximations can occur, is given by 
\begin{equation}\label{eq:radial_bound}
    \overline{\rho} = r_e + r_o + \frac{d_{o}}{2}\big(N_{o} - 1\big) + \frac{d_{e}}{2}\big(N_{e} -1\big),
\end{equation}
see \cite{Tolksdorf_POC_2024}. We use a polar coordinate frame because that allows us to restrict the maximum collision distance (\ref{eq:radial_bound}) solely in the radial coordinate, hence, we define a change of coordinates as 
\begin{equation}\label{eq:polar_transform}
\begin{split}
   & CT: \mathbb{R} \times [0, 2\pi) \rightarrow \mathbb{R}^2,\\
   & (\rho, \phi) \mapsto (x, y) = (\rho \cos{\phi}, \rho \sin{\phi}).
\end{split}
\end{equation} 
Then, for all angles $\phi \in [0, 2\pi)$, collisions are possible for $\rho \in [0, \overline{\rho}]$. We denote $\mathbf{y}_{p} := (\rho, \phi, \theta)$ a polar representation of the configuration, for which it holds that $\mathbf{y}_{p}= (\sqrt{x^2 + y^2}, \text{atan2}(y, x), \theta)$, where `$\text{atan2}$' designates the two-argument arctangent function. Recall that $p_{\mathbf{y}}(x, y, \theta; \mathbf{w}_{\mathbf{y}}) = p_{\mathbf{x}}(x, y; \mathbf{w}_{\mathbf{x}})p_{\theta}(\theta; \mathbf{w}_\theta)$, where we apply the same change of coordinates to $p_{\mathbf{x}}$, giving $p_{\phi, \rho}$, and, hence, $p_{\mathbf{y}}(x, y, \theta; \mathbf{w}_{\mathbf{y}}) = p_{\phi, \rho}(\phi, \rho; \mathbf{w}_{\phi, \rho})p_{\theta}(\theta; \mathbf{w}_\theta)$ with the parameterization $\mathbf{w}_{\phi, \rho}$ for the position.\\
We denote by $\mathbb{I}^{\theta}_{j,l}(\phi, \rho) = [\underline{\theta}_{j,l}(\phi, \rho), \overline{\theta}_{j,l}(\phi, \rho)]$ an intersection angle interval, i.e., an interval for the heading angle $\theta$ for which the object circle $l$ intersects with the ego circle $j$ for a specific position $(\phi, \rho) \in [0, 2\pi] \times [0, \overline{\rho}]$. For the detailed derivation of expressions for $\mathbb{I}_{j,l}^{\theta}(\phi, \rho)$, we refer to \cite{Tolksdorf_POC_2024}. With all $N_e N_o$ intersection angle intervals, for a specific $(\phi, \rho) \in [0, 2\pi) \times [0, \overline{\rho}]$, the region of integration of the PDF for the object's heading angle $\theta$ is given by 
\begin{equation}\label{eq:collision_set}
    \breve{\mathbb{I}}^{\theta}(\phi, \rho) := \bigcup_{j = 1}^{N_e}\bigcup_{l = 1}^{N_o} \mathbb{I}^{\theta}_{j,l}(\phi, \rho) = \bigcup_{i=1}^{N_I}\breve{\mathbb{I}}^{\theta}_{i}(\phi, \rho).
\end{equation}
Note that, in (\ref{eq:collision_set}), several intervals intervals $\mathbb{I}^{\theta}_{j,l}(\phi, \rho)$ may overlap, which is illustrated in Example \ref{ex:overlap} below. The right-most expression in (\ref{eq:collision_set}) therefore denotes that there might be up to $N_I$ disjoint intervals $\breve{\mathbb{I}}^{\theta}_{i}(\phi, \rho)$.
Thus, the set of all polar object configurations leading to a collision with the ego vehicle is given by 
\begin{equation}\label{eq:Acir}
    \mathcal{A}_{cir, p} = \{(\phi, \rho, \theta) \in \mathcal{C}_p\! \mid \! \phi \in [0, 2\pi), \rho \in [0, \overline{\rho}], \theta \in \breve{\mathbb{I}}^{\theta}(\phi, \rho) \},
\end{equation}
where $\mathcal{C}_p$ is the polar transformation of the configuration space, i.e., $\mathcal{C}_p := CT^{-1}(\mathbb{R}^2) \times \mathbb{R}$. Note that the same transformation can also be applied to $\mathcal{A}_{rec}$, giving $\mathcal{A}_{rec,p}$. With these considerations, (\ref{eq:general_POC}) can be over-approximated, i.e., $\mathcal{A}_{cir, p} \supset \mathcal{A}_{rec,p}$, as
\begin{equation}\label{eq:poc_overapproximation}
\begin{split}
    &\mathbb{P}\{\mathbf{y}_{p} \in \mathcal{A}_{cir, p}\} = \\
    &\int_0^{2\pi}\int_0^{\overline{\rho}} p_{\phi, \rho}(\cmbphi, \cmbrho; \mathbf{w}_{\phi, \rho} )
    \left[  \sum_{i=1}^{N_{I}} \int_{\breve{\mathbb{I}}^{\theta}_{i}(\cmbphi, \cmbrho)} p_{\theta}(\cmbtheta; \mathbf{w}_{\theta})\text{d}\cmbtheta \right]\text{d}\cmbrho \text{d}\cmbphi.
\end{split}
\end{equation} 
\begin{example}\label{ex:overlap}
     In Figure \ref{fig:severity_intervals}, four exemplary cases are provided by rotating the object at the same position $(\phi, \rho)$. Figure \ref{fig:severity_intervals}(a): No object circle is intersecting with any ego circle if the object is rotated to $\theta_{1}$. Figure \ref{fig:severity_intervals}(b): The front object circle is intersecting with the front ego circle for $\theta_{2}$. In fact, the object's front circle is intersecting with the ego front circle for all $\theta \in \mathbb{I}^{\theta}_{1,1}(\phi, \rho)$, depicted by the orange heading angle range. Figure \ref{fig:severity_intervals}(c): When the object's heading angle is increased to $\theta_{3}$, the object's front circle is not only intersecting with the ego's front circle, but also the ego's center circle. The range for which that happens is the intersection $\mathbb{I}^{\theta}_{1,1}(\phi, \rho) \cap \mathbb{I}^{\theta}_{2,1}(\phi, \rho)$. Figure \ref{fig:severity_intervals}(d): Lastly, when the object's heading is further increased to $\theta_{4}$, the object's front circle is only intersecting with the ego's center circle within the heading angle range of $\mathbb{I}^{\theta}_{2,1}(\phi, \rho)$.
\end{example}
The next step is to extend the probability of collision toward a risk measure by also considering the collision severity, see  (\ref{eq:risk_general}). Note that the circles of one vehicle can overlap (see the circular placement per vehicle in Figure \ref{fig:severity_intervals}). Hence, there is one or more areas (depending on the respective number of circles $N_e$ and $N_o$) with two severity functions, essentially just creating another collision area which the function $f_s$ must account for. Combining all severity functions $s_{j,l}$ within (\ref{eq:total_severity}) and using the total severity function and the characterization for the probability of collision  (\ref{eq:poc_overapproximation}), the risk in (\ref{eq:risk_general}) can be estimated (in fact, over-approximated) with
\begin{equation}\label{eq:ego_risk_full}
    \begin{split}
        &R^{cir}(v_e; \mathbf{w})=
        \int_0^{2\pi}\!\int_0^{\overline{\rho}}\!p_{\phi, \rho}(\cmbphi, \cmbrho; \mathbf{w}_{\phi, \rho})\Bigg[\sum_{i=1}^{N_{I}}\\
        & \int_{\breve{\mathbb{I}}^{\theta}_{i}(\cmbphi, \cmbrho)} \int_{\mathcal{V}_o}\!p_{\theta}(\cmbtheta; \mathbf{w}_{\theta})p_{v_o}(\cmbv_o; \mathbf{w}_{v_o})s(\cmbtheta, v_e, \cmbv_o)\text{d}\cmbv_o \text{d} \cmbtheta\Bigg]\text{d}\cmbrho \text{d}\cmbphi,
    \end{split}
\end{equation}
where $\mathbf{w} = (\mathbf{w}_{\phi, \rho}, \mathbf{w}_{\theta}, \mathbf{w}_{v_o})$.
\subsection{Combining Severity Functions}
In (\ref{eq:ego_risk_full}), the total severity function $s$ is integrated over $\theta$ for $N_I$ disjoint intersection angle intervals $\breve{\mathbb{I}}^{\theta}_{i}(\phi, \rho)$. However, our individual severity functions $s_{j,l}$ are defined on a circle-to-circle basis. Due to the problem of overlap (see Example \ref{ex:overlap}), integrating $\theta$ over all $N_eN_o$ intervals $\mathbb{I}^{\theta}_{j,l}(\phi, \rho)$ would result in a much larger over-approximation as the areas within an overlap are accounted for multiple times. In (\ref{eq:poc_overapproximation}), this is resolved by taking the union over all intersection angle intervals $\mathbb{I}^{\theta}_{j,l}(\phi, \rho)$ in (\ref{eq:collision_set}), since the information about which circle-to-circle pair is colliding is not needed for the estimation of only the probability of collision. However, such information is now of importance for the severity $f_s$, since otherwise one would not be able to differentiate between collision constellations.
Hence, we require a method to combine the individual severity functions $s_{j,l}$ into a total severity value $s$, that can be used in (\ref{eq:ego_risk_full}), such that the domain of $\theta$ remains disjoint but the information about the collision constellation is not discarded.\\
We approach deriving a total severity value by using a sum of the individual severities, which is averaged whenever multiple intersection angle intervals $\mathbb{I}^{\theta}_{j,l}(\phi, \rho)$ intersect. Therefore, we integrate the individual severity function $s_{j,l}$ over the object's velocity $v_o$, as this variable is not considered in collision determination, giving the conditional expected individual severity with respect to the object's velocity:
\begin{equation}\label{eq:expected_severity}
    \mathbb{E}[s_{j,l}(\theta, v_e, v_o)\mid v_o] = \int_{\mathcal{V}_o}p_{v_o}(\cmbv_o; \mathbf{w}_{v_o})s_{j,l}(\theta, v_e, \cmbv_o)\text{d}\cmbv_o.
\end{equation}
Using $\mathbb{E}[s_{j,l}(\theta, v_e, v_o) \mid v_o]$, we construct the sum of average severities by following two rules. First, if a subinterval of a particular $\mathbb{I}^{\theta}_{j,l}(\phi, \rho)$ is disjoint from all other subintervals of $\mathbb{I}^{\theta}_{u,v}(\phi, \rho)$, with $(j,l)\neq (u,v)$, then it's severity is $s_{j,l}$. Second, if an intersection among all $N_eN_o$ intervals $\mathbb{I}^{\theta}_{j,l}(\phi, \rho)$ is found, the severity for the intersecting part is the average of all severity functions in that intersection. Thus, the severity will be given by a sum of disjoint and intersecting interval parts. We illustrate our approach to construct the sum of average severities with an example.\begin{figure}[h]
\begin{center}
\includegraphics[width=8.4cm]{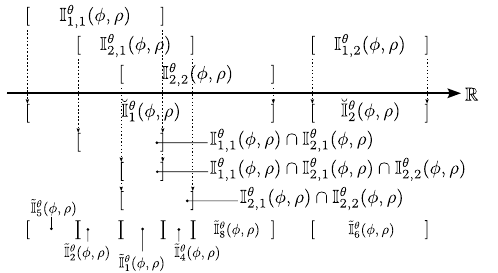}  
\caption{Illustration of Example \ref{ex:intersection_intervals}: Construction of disjoint intersection angle intervals.}
\label{fig:intersection_intervals}
\end{center}
\end{figure}
\begin{example}\label{ex:intersection_intervals}
Figure \ref{fig:intersection_intervals} shows the intersection intervals in $\theta$ for the case of an ego and object covering of two circles each. The derived (overlapping) intersection intervals $\mathbb{I}^{\theta}_{j,l}(\phi, \rho)$ are depicted above the real number line. In the row directly below the real number line there are two intervals $\breve{\mathbb{I}}^{\theta}_{i}(\phi, \rho)$, resulting form the union over all $\mathbb{I}^{\theta}_{j,l}(\phi, \rho)$ see (\ref{eq:collision_set}). Clearly, some of the intervals $\mathbb{I}^{\theta}_{j,l}(\phi, \rho)$ above overlap. Intersections of both two and three intervals are present, depicted in rows two to four below the real number line. The disjoint intervals in the last row are derived in (\ref{eq:disjoint_remainder}), of which the derivation will be discussed in the following exposition.
\end{example} 
\begin{table*}
    \centering
    \caption{Application of Algorithm \ref{alg:weigthed_severity} to the Example \ref{ex:intersection_intervals} depicted in Figure \ref{fig:intersection_intervals}.}
    \label{tab:intersection_example}
    \begin{tabularx}{\linewidth}{ C | C | C | C | C | C | C | C | C | C}
        \toprule
       m& $\tau$ & \multicolumn{2}{c|}{$\mathcal{L}_m$} & \multicolumn{2}{c|}{$\mathbb{I}_m^{\theta}(\phi, \rho)$} & \multicolumn{2}{c|}{$\tilde{\mathbb{I}}^\theta_m(\phi, \rho)$} & \multicolumn{2}{c}{$\mathbb{E}[s_{m}]$}\\
        \toprule
        \midrule
            1 & 3 & \multicolumn{2}{c|}{$\{ (1,1), (2,1), (2,2)\}$} & \multicolumn{2}{c|}{$[\underline{\theta}_{2,2}(\phi, \rho), \overline{\theta}_{1,1}(\phi, \rho)]$} & \multicolumn{2}{c|}{ $\mathbb{I}_1^{\theta}(\phi, \rho)$ } & \multicolumn{2}{c}{$\frac{1}{3}(\mathbb{E}[s_{1,1}] + \mathbb{E}[s_{2,1}] + \mathbb{E}[s_{2,2}])$}\\
            \midrule
            2 & 2 & \multicolumn{2}{c|}{$\{ (1,1), (2,1)\}$}  & \multicolumn{2}{c|}{$[\underline{\theta}_{2,1}(\phi, \rho), \overline{\theta}_{1,1}(\phi, \rho)]$} & \multicolumn{2}{c|}{$\mathbb{I}_2^{\theta}(\phi, \rho)\setminus\tilde{\mathbb{I}}_1^\theta(\phi, \rho)$} & \multicolumn{2}{c}{ $\frac{1}{2}(\mathbb{E}[s_{1,1}] + \mathbb{E}[s_{2,1}])$ }\\
            \midrule
            3 & 2 & \multicolumn{2}{c|}{$\{ (1,1), (2,2)\}$}  & \multicolumn{2}{c|}{ $[\underline{\theta}_{2,2}(\phi, \rho), \overline{\theta}_{1,1}(\phi, \rho)]$ } & \multicolumn{2}{c|}{ $\mathbb{I}_3^{\theta} (\phi, \rho)\setminus\bigcup_{u<m}\tilde{\mathbb{I}}^\theta_u(\phi, \rho)$} & \multicolumn{2}{c}{$\frac{1}{2}(\mathbb{E}[s_{1,1}] + \mathbb{E}[s_{2,2}])$}\\
            \midrule
            4 & 2 & \multicolumn{2}{c|}{$\{ (2,1), (2,2)\}$}  & \multicolumn{2}{c|}{$[\underline{\theta}_{2,2}(\phi, \rho), \overline{\theta}_{2,1}(\phi, \rho)]$} & \multicolumn{2}{c|}{$\mathbb{I}_4^{\theta} (\phi, \rho)\setminus\bigcup_{u<m}\tilde{\mathbb{I}}^\theta_u(\phi, \rho)$} & \multicolumn{2}{c}{$\frac{1}{2}(\mathbb{E}[s_{2,1}] + \mathbb{E}[s_{2,2}])$}\\
            \midrule
            5 & 1 & \multicolumn{2}{c|}{$\{(1,1)\}$}  & \multicolumn{2}{c|}{$[\underline{\theta}_{1,1}(\phi, \rho), \overline{\theta}_{1,1}(\phi, \rho)]$} & \multicolumn{2}{c|}{$\mathbb{I}_5^{\theta} (\phi, \rho)\setminus\bigcup_{u<m}\tilde{\mathbb{I}}^\theta_u(\phi, \rho)$} & \multicolumn{2}{c}{$\mathbb{E}[s_{1,1}]$}\\
            \midrule
            $\vdots$ & $\vdots$ & \multicolumn{2}{c|}{$\vdots$}  & \multicolumn{2}{c|}{$\vdots$} & \multicolumn{2}{c|}{$\vdots$} & \multicolumn{2}{c}{$\vdots$}\\
            \midrule            
            8 & 1 & \multicolumn{2}{c|}{$\{(2,2)\}$}  & \multicolumn{2}{c|}{$[\underline{\theta}_{2,2}(\phi, \rho), \overline{\theta}_{2,2}(\phi, \rho)]$} & \multicolumn{2}{c|}{$\mathbb{I}_8^{\theta} (\phi, \rho)\setminus\bigcup_{u<m}\tilde{\mathbb{I}}^\theta_u(\phi, \rho)$} & \multicolumn{2}{c}{$\mathbb{E}[s_{2,2}]$}
    \end{tabularx}
\end{table*}
To calculate the sum of average severities, three operations are necessary. First, one must determine all intersections among all intervals $\mathbb{I}^{\theta}_{j,l}(\phi, \rho)$. Here, we distinguish the order of intersection index $\tau$, where $\tau$ indicates the number of intervals within the resulting disjoint interval, hence, $0 <\tau \leq N_eN_o$ (e.g, $\tau = 1$ for $\tilde{\mathbb{I}}_5^{\theta}(\phi, \rho)$, or $\tau = 3$ for $\tilde{\mathbb{I}}_1^{\theta}(\phi, \rho)$ in Figure \ref{fig:intersection_intervals}, last row). Second, one must ensure that each intersection is not included in another intersection. And, third, the average severity of each intersection must be calculated. To implement this procedure, we design an algorithm that starts identifying the intersection of the highest order $\tau$ first and then identifies the remaining intersections in descending order, as lower-order intersections may already have been accounted for in a higher-order intersection. The algorithm is iterative. Every iteration ends when a new intersection or disjoint sub-inteval is found. Now, let $m$ denote the current algorithm iteration. The three operations described above are performed as follows: first, we construct a set of tuples of indices $\mathcal{L}_m$ preserving the information of which circle-to-circle pairs are intersecting. For example, $\mathcal{L}_m = \{(j,l), (u,v)\}$ contains the indices of $\mathbb{I}^{\theta}_{j,l}(\phi, \rho)$ and $\mathbb{I}^{\theta}_{u,v}(\phi, \rho)$. With $\mathcal{L}_m$, we can identify the intersections of $\mathbb{I}^{\theta}_{j,l}(\phi, \rho)$ as follows: 
\begin{equation}\label{eq:intersections}
    \mathbb{I}_m^{\theta}(\phi, \rho) = \bigcap_{(j,l) \in \mathcal{L}_m }\mathbb{I}^\theta_{j,l}(\phi, \rho).
\end{equation}
Given $\mathbb{I}_m^{\theta}(\phi, \rho)$, we execute the second operation, i.e., ensuring disjointedness of the intersection with any other intersections,
\begin{equation}\label{eq:disjoint_remainder}
    \tilde{\mathbb{I}}^\theta_m(\phi, \rho) = 
    \mathbb{I}_m^{\theta}(\phi, \rho)\setminus\bigcup_{u \in \{1, 2, \dots, m -1\}}\tilde{\mathbb{I}}^\theta_{u}(\phi, \rho).
\end{equation}
For the third step, we average the severities by the cardinality of $\mathcal{L}_m$, i.e., $|\mathcal{L}_m| = \tau$, giving the average severity for that iteration $m$, which, when considering the expected value of that, reads
\begin{equation}\label{eq:weighted_sum}
\begin{split}
    \mathbb{E}[s_{m}(\theta, v_e, v_o) \mid v_o] =
    \frac{1}{|\mathcal{L}_m|} \sum_{(j,l) \in \mathcal{L}_m}\mathbb{E}[s_{j,l}(\theta, v_e, v_o)\mid v_o],
\end{split}
\end{equation}
where $\mathbb{E}[s_{j,l}(\theta, v_e, v_o)\mid v_o]$ is given by (\ref{eq:expected_severity}). With (\ref{eq:expected_severity}) and (\ref{eq:weighted_sum}), we can express the inner integrals in (\ref{eq:ego_risk_full}) with respect to $v_o, \theta$ as
\begin{equation}\label{eq:inner_integral}
\begin{split}
\sum_{m=1}^{N_{m}} \int_{\tilde{\mathbb{I}}^{\theta}_{m}(\cmbphi, \cmbrho)} \!p_{\theta}(\cmbtheta; \mathbf{w}_{\theta})\mathbb{E}&[s_m(\cmbtheta, v_e, v_o) \mid v_o] \text{d} \cmbtheta \\
&=: \mathbb{E}[s(\theta, v_e, v_o) \mid \theta, v_o, \phi, \rho \;].
\end{split}
\end{equation}
\begin{algorithm}
\caption{Expected Average Severity}\label{alg:weigthed_severity}
\begin{algorithmic}[1]
\Require $\mathbb{I}^{\theta}_{j,l}(\phi, \rho)$
\State $\tau \gets N_eN_o$
\State $m \gets 1$
\State $stop \gets \texttt{False}$
\State $\mathbb{E}[s] \gets 0$
\While{$stop$ == \texttt{False}}
	\State $\mathcal{L}_m, itr \gets \text{permuteIndices}(\tau, m)$
    \State $\mathbb{I}_m^{\theta}(\phi, \rho) \gets \bigcap_{(j,l) \in \mathcal{L}_m}\mathbb{I}^\theta_{j,l}(\phi, \rho)$ \Comment{see (\ref{eq:intersections})}
    \If {$\mathbb{I}_m^{\theta}(\phi, \rho) \neq \emptyset$}
    \State $\tilde{\mathbb{I}}^\theta_m(\phi, \rho) \gets 
    \mathbb{I}_m^{\theta}(\phi, \rho)\setminus\bigcup_{u < m}\tilde{\mathbb{I}}^\theta_{u}(\phi, \rho)$ \Comment{see (\ref{eq:disjoint_remainder})}
    \If{$\tilde{\mathbb{I}}^\theta_m(\phi, \rho) \neq \emptyset$}
    \State $\mathbb{E}[s_m] \gets \frac{1}{|\mathcal{L}_m|}  \sum_{(j,l) \in \mathcal{L}_m}\mathbb{E}[s_{j,l}]$ \Comment{see (\ref{eq:weighted_sum})}
    \State $\mathbb{E}[s] \gets \mathbb{E}[s] + \int_{\tilde{\mathbb{I}}^{\theta}_{m}(\cmbphi, \cmbrho)} \!p_{\theta}(\cmbtheta; \mathbf{w}_{\theta})\mathbb{E}[s_m] \text{d}\cmbtheta $
    \EndIf
    \State $m = m + 1$
    \EndIf
    \If{$itr == \texttt{True}$}
        \State $\tau \gets \tau -1$
        \If{$\tau < 1$}
        \State $stop \gets \texttt{True}$ \Comment{stop while loop}
        \EndIf
    \EndIf
\EndWhile\\
\Return $\mathbb{E}[s]$
\end{algorithmic}
\end{algorithm}
We integrated operations (\ref{eq:intersections}) - (\ref{eq:inner_integral}) in Algorithm \ref{alg:weigthed_severity}, which iterates over all intervals in descending order of intersections $\tau$. Note that we abbreviated the expected values in Algorithm \ref{alg:weigthed_severity} and Table \ref{tab:intersection_example} for conciseness. Further, the function `permIndicies($\tau, m$)' in line 6 of Algorithm \ref{alg:weigthed_severity} permutes the circle-to-circle indices $j, l$, returns the set $\mathcal{L}_m$, and assigns $itr$ to check whether all possible permutations within the intersection order $\tau$ have been calculated. In Table \ref{tab:intersection_example}, we recursively apply Algorithm \ref{alg:weigthed_severity} to Example~\ref{ex:intersection_intervals}. 
Finally, the severity term in (\ref{eq:ego_risk_full}) is estimated with (\ref{eq:expected_severity}) - (\ref{eq:weighted_sum}) and the result is re-introduced to (\ref{eq:ego_risk_full}) with (\ref{eq:inner_integral}), which gives the risk as 
\begin{equation}\label{eq:ego_risk}
    \begin{split}
        &R^{cir}(v_e; \mathbf{w})=
        \int_0^{2\pi}\!\int_0^{\overline{\rho}}\!p_{\phi, \rho}(\cmbphi, \cmbrho; \mathbf{w}_{\phi, \rho})\\
        &  \Bigg[\sum_{m=1}^{N_{m}}\int_{\tilde{\mathbb{I}}^{\theta}_{m}(\cmbphi, \cmbrho)} \!p_{\theta}(\cmbtheta; \mathbf{w}_{\theta})\mathbb{E}[s_m(\cmbtheta, v_e, v_o) \mid v_o] \text{d} \cmbtheta\Bigg]\text{d}\cmbrho \text{d}\cmbphi.
    \end{split}
\end{equation}
\begin{remark}
In~\cite{Tolksdorf_POC_2024}, the proof of Theorem 1 guarantees that (\ref{eq:poc_overapproximation}) is an over-approximation of the true probability of collision under the conditions that the intervals $\breve{\mathbb{I}}^{\theta}_{i}(\phi, \rho)$ are disjoint and the circular placement fully encloses the rectangular shapes of both vehicles. We note that both of these conditions still hold when using $\tilde{\mathbb{I}}^{\theta}_{m}(\phi, \rho)$ instead of $\breve{\mathbb{I}}^{\theta}_{i}(\phi, \rho)$ as the circular placement remains the same and the intervals $\tilde{\mathbb{I}}^{\theta}_{m}(\phi, \rho)$ are disjoint for all $m \in \{1, 2, \dots, N_m\}$.
\end{remark}
\section{Risk Estimation for Gaussian Uncertainties and a Kinematic Severity Model}\label{sec:risk_for_gaussian_uncertainties}
This section evaluates the risk from (\ref{eq:ego_risk}) for Gaussian uncertainties and a kinematic collision severity model, as the previous sections neither details the type of probability density functions nor the severity functions $s_{j,l}$. We use Gaussians as those are often used (see, e.g.,~\cite{dixit2019trajectory, volz2015stochastic, patil2012estimating, ploeg2022long, altendorfer2021new, du2011probabilistic}) to model kinematic uncertainties in motion planning, hence, we model $x,y, v_o$ to be Gaussian random variables and assign a wrapped Gaussian distribution to $\theta$. As a severity function, we will use a kinematic energy model that can accommodate different collision constellations, as the impact velocities and masses form a suitable predictor of collision severity (see, e.g., \cite{tolouei2013vehicle, jurewicz2016exploration}). 
\subsection{Gaussian Uncertainty}
Although a relative heading angle $\theta$ is defined on $\mathbb{R}$, to any observer it appears periodic on the interval $[0, 2\pi)$. To account for that, we use a wrapped normal distribution to assign a Gaussian-like probability density function to the relative heading angle. With $\mu_{\theta}, \sigma^2_{\theta}$ being the mean and variance of $\theta$ composed in $\mathbf{w}_{\theta} = (\mu_\theta, \sigma_\theta)$, let the wrapped-Gaussian for the relative heading angle be given by 
\begin{equation}\label{eq:inf_wrapped_gauss}
    p_{\theta}(\theta; \mathbf{w}_{\theta})\!=\!\frac{1}{\sqrt{2\pi} \sigma_{\theta}}\!\sum_{\beta = -\infty}^{\infty}\!\exp{\left[ -\frac{(\theta\!+\!2\pi \beta\! -\!\mu_{\theta})^2}{2\sigma_{\theta}^2} \right]}.
\end{equation}
As shown in \cite{kurz_wrapped_gauss}, the infinite sum in (\ref{eq:inf_wrapped_gauss}) can be tightly approximated by truncating it within $\beta \in \mathbb{Z}_{-N_\beta}^{N_\beta}$, when $N_\beta \geq 3$, leaving $2N_\beta + 1$ summands\footnote{In fact, the accuracy of this approximation also depends on $\sigma_{\theta}$ and the error $e_\beta$ (i.e., the absolute difference between the truncated and the true value in (\ref{eq:inf_wrapped_gauss})) grows when $\sigma_{\theta}$ increases. For example, let $N_\beta = 3$ and $\sigma_{\theta} = 1$, then $e_\beta < 10^{-15}$, if $\sigma_{\theta}$ = 5; then $e_\beta < 10^{-6}$ \cite{kurz_wrapped_gauss}. Generally, for realistic values of $\sigma_{\theta}$, i.e., $\sigma_{\theta} \leq \pi$, the error is insignificant, especially when comparing it to the precision of numerical integration, which we will later set to three digits.}.\\
Given the polar coordinate transformation (\ref{eq:polar_transform}), we retrieve the bivariate-Gaussian for the positional coordinates $\mathbf{x}$ as a function of the polar coordinates $\phi$ and $\rho$ as follows:
\begin{equation}\label{eq:polar_gauss}
\begin{split}
    & p_{\phi, \rho}(\phi, \rho; \mathbf{w}_{\phi, \rho}) = \frac{\rho}{2 \pi \sigma_{x}\sigma_{y}} \text{exp} \Bigg[ \frac{(\rho \cos{(\phi)} - \mu_{x})^2}{-2\sigma_{x}^2} \\
    & -\frac{(\rho \sin{(\phi)} -  \mu_{y})^2}{2\sigma_{y}^2}  \Bigg],
\end{split}
\end{equation} 
where $\mu_{x}, \mu_{y}, \sigma^2_{x}, \sigma^2_{y}$ are the means and variances of $x, y$, composed in $\mathbf{w}_{\phi, \rho} = ((\mu_x, \sigma_x),(\mu_y, \sigma_y))^T$ and we note that the integral over (\ref{eq:polar_gauss}) in (\ref{eq:ego_risk}) requires numerical integration (see \cite{tolksdorf2024}).\\
Lastly, the uncertainty in the velocity is given by a univariate Gaussian with mean $\mu_{v_o}$ and variance $\sigma_{v_o}^2$, where the cumulative distribution function with $\mathcal{V}_o = [\underline{v}_o, \overline{v}_o]$ reads
\begin{equation}\label{eq:analytic_univariant_gauss}
\begin{split}
    &F_v(\mathcal{V}_o; \mathbf{w}_{v_o}) := \int_{\mathcal{V}_o}p_{v_o}(\cmbv_o; \mathbf{w}_{v_o})\text{d}\cmbv_o\\
    &=\frac{1}{2}\left[\text{erf}\left(\frac{\overline{v}_o- \mu_{v_o}}{\sigma_{v_o} \sqrt{2}}\right)-\text{erf}\left(\frac{\underline{v}_o- \mu_{v_o}}{\sigma_{v_o} \sqrt{2}}\right)\right],
\end{split}
\end{equation}
where $\text{erf}(a)$ denotes the error function of $a$. Note that~(\ref{eq:analytic_univariant_gauss}) will be a useful expression to derive the conditional expected severity next.
\subsection{Severity Function}
Whether an analytical expression for (\ref{eq:ego_risk}) can be found depends on the form of the severity function $s_{j,l}$ and the uncertainty characterization in terms of the PDFs. The directionality of each actor's velocity is, however, challenging to consider when using Gaussian uncertainty, as an analytical solution to (\ref{eq:inner_integral}) may not be obtainable. The proposed multi-circular model, however, can accommodate different severity functions for different collision constellations regardless of the heading angle. We, therefore, propose a kinetic energy model which weighs collisions differently, depending on the specific circle-to-circle collision. Hence, we have
\begin{equation}\label{eq:severity}
    s_{j,l}(v_e, v_o) = \frac{w_{j,l}m_em_o}{2(m_e + m_o)}\cdot\begin{cases}
    (v^2_e + v^2_o) \text{ head-on coll.,}\\
    v_e^2 \text{ ego-to-obj. side coll.,}\\
    v_o^2 \text{ obj.-to-ego side coll.,}\\
    (v^2_e - v^2_o) \text{ ego rear-end coll.,}\\
    (v^2_o - v^2_e) \text{ obj. rear-end coll.,}
\end{cases}
\end{equation}
with $m_e, m_o$ being the ego and object mass, respectively. The cases in (\ref{eq:severity}) are assigned based on the specific circle-to-circle collision pairs\footnote{Assigning the collision cases is a design choice. Suppose a three circle approximation of each actor, then the five cases in (\ref{eq:severity}) must be assigned to nine possible circle-to-circle collision pairs. For example, \textit{ego rear-end collision} denotes a case when the ego's front circle collides to the object's rear circle.}. In addition, $w_{j,l}$ denotes a positive scalar weighting factor that accounts for the severity of the $j$-th ego circle colliding with the $l$-th object circle. With the severity function and the integral of the univariate Gaussian in the velocity (\ref{eq:analytic_univariant_gauss}), we can derive the conditional expected value of (\ref{eq:severity}), with respect to the object's velocity $v_o$, by using two integrals, $I_e$ and $I_o$, respectively. The first integral $I_e$ represents the integral of the univariate Gaussian $p_{v_o}$ multiplied with the constant $v_e^2$, see (\ref{eq:analytic_univariant_gauss}). The second integral, $I_o$, is a product of $v_o^2$ and $p_{v_o}$, where the solution is known, e.g., see \cite{owen1980table}. Hence, the conditional expected circle-to-circle severity is given by:
\begin{equation}\label{eq:analytic_severity}
\begin{split}
    &\mathbb{E}[s_{j,l}(v_e, v_o) \mid v_o] = 
    c_1\cdot \begin{cases}(I_e + I_o)\text{ head-on coll.,}\\
    I_o \text{ ego-to-obj. side coll.},\\
    I_e \text{ obj.-to-ego side coll.}, \\
    (I_e - I_o) \text{ ego rear-end coll.,}\\
    (I_o - I_e) \text{ obj rear-end coll.,}
    \end{cases}\\
    &\text{where: }
    I_e = \frac{v_e^2}{2}\left[\text{erf}\left( \frac{\overline{v}_o - \mu_{v_o}}{\sigma_{v_o} \sqrt{2}}\right) - \text{erf}\left( \frac{\underline{v}_o - \mu_{v_o}}{\sigma_{v_o} \sqrt{2}}\right) \right],\\
    & I_o = \frac{1}{\sigma_{v_o} c_3^3}\Bigg[ \frac{c_2^2 + 1}{2} \left( \text{erf}\left(\frac{c_2 + c_3\overline{v}_o}{\sqrt{2}} \right) - \text{erf}\left(\frac{c_2 + c_3\underline{v}_o}{\sqrt{2}} \right)\right)  \\
    & \quad + \frac{1}{\sqrt{2\pi}}\Bigg((c_2 - c_3\overline{v}_o) \exp{\left(\frac{(c_2+c_3\overline{v}_o)^2}{-2} \right)}\\
    & \quad - (c_2 - c_3\underline{v}_o)\exp{\left(\frac{(c_2+c_3\underline{v}_o)^2}{-2} \right)}\Bigg)\Bigg],\\
    & c_1 = \frac{w_{j,l}m_em_o}{2(m_e + m_o)}, \quad c_2 = \frac{-\mu_{v_o}}{\sigma_{v_o}}, \quad c_3 = \frac{1}{\sigma_{v_o}}.
\end{split}
\end{equation}
Note that we substitute (\ref{eq:analytic_severity}) (albeit without the argument of $\theta$) in (\ref{eq:weighted_sum}) to retrieve $\mathbb{E}[s(v_e, v_o)\mid v_o]$ with Algorithm \ref{alg:weigthed_severity}.

\subsection{Risk Estimation}
While (\ref{eq:severity}) does not depend on $\theta$ (which may not be case in general, see (\ref{eq:inner_integral})), it is still part of the uncertainty in the configuration. However, by assumption, $\theta$ is independent of the positional coordinates $\phi, \rho$. Hence, we can solve the integral of $\theta$ in (\ref{eq:ego_risk}) irrespective of the severity term and the the positional coordinates $\phi, \rho$.
Therefore, we analytically integrate (\ref{eq:inf_wrapped_gauss}) over $\bigcup_{m = 1}^{N_m}\tilde{\mathbb{I}}_{m}^{\theta}(\phi, \rho)$, given the approximation $-N_\beta \leq \beta \leq N_\beta$ in the sum in (\ref{eq:inf_wrapped_gauss}). Hence, for the conditional expected average severity (\ref{eq:inner_integral}) we have
\begin{equation}\label{eq:analytic_warpped_gauss}
\begin{split}
    & \mathbb{E}[s(v_e, v_o) \mid \theta, v_o, \phi, \rho \;] = \\
    &\sum_{m = 1}^{N_m}\Bigg\{\sum_{\beta=-N_\beta}^{N_\beta}\Bigg[\text{erf}\left( \frac{\overline{\theta}_{m}(\phi, \rho)\!-\!\mu_{\theta}\!+\!2\pi \beta}{\sigma_{\theta} \sqrt{2}}\right)\\
    & - \! \text{erf}\left( \frac{\underline{\theta}_{m}(\phi, \rho)\!-\!\mu_{\theta}\!+\!2\pi \beta}{\sigma_{\theta} \sqrt{2}}\right)\!\Bigg]\mathbb{E}[s_m(v_e, v_o) \mid v_o]\Bigg\}.
\end{split}
\end{equation}
With (\ref{eq:analytic_warpped_gauss}), the risk from (\ref{eq:ego_risk}) is given by 
\begin{equation}\label{eq:ego_risk_short}
\begin{split}
    R^{cir}(v_e; \mathbf{w}) = \int_0^{2\pi}\int_0^{\overline{\rho}}&p_{\phi, \rho}(\cmbphi, \cmbrho; \mathbf{w}_{\phi, \rho})\\
    &
    \mathbb{E}[s(v_e, v_o) \mid \theta, v_o, \cmbphi, \cmbrho \;] \text{d}\cmbrho \text{d}\cmbphi,
\end{split}\end{equation}
where we solve the remaining two-dimensional integral numerically\footnote{
    Note that a computationally efficient algorithm to compute the right-hand side of (\ref{eq:ego_risk_short}) without the severity term is presented in \cite{Tolksdorf_POC_2024}.
} using the expressions in (\ref{eq:analytic_warpped_gauss}) and (\ref{eq:polar_gauss})\footnote{The programming code can be accessed online here: \url{https://github.com/Tolksdorf/Risk-Estimation.git}.}.
\section{Numerical Case Studies}\label{sec:numerical_case_studies}
In this section, we present a numerical case study to demonstrate the proposed methodology for risk estimation in different collision scenarios and assess its computational performance.
\subsection{Collision Scenarios} 
A key contribution of this paper is that the proposed risk function features a severity model that takes different impact constellations into account by associating them with different severity values. Therefore, we design five exemplary test cases that demonstrate the efficacy of the risk function in frequently occurring real-world accidents, i.e., head-on, rear-end, and side collisions with varying impact points. Throughout all scenarios, the motion of each vehicle is (assumed to be) unaffected by the motion of the other,  allowing both vehicles to travel through each other. Note that the initial configurations, velocities, and uncertainties are provided in \autoref{tab:Case_Study_Parameters}, where the configurations are in a global coordinate frame. Therefore, we distinguish the ego's and object's variables by $e$ and $o$ subscripts, respectively.
All velocities are kept constant for each test case. Cases (I) and (II) examine the risk function in a head-on and a rear-end collision scenario, respectively, where the object is traveling along the ego's longitudinal axis, see \autoref{fig:NumCaseHeadOnRearEnd}.
Cases (III), (IV), and (V) consider side collisions, where the object collides with the ego at a $90\,\text{deg}$ angle at three different impact points, i.e., front, center, and rear, respectively (see \autoref{fig:NumCaseSides}). Case (III) focuses on a collision with the front of the ego, whilst Case (IV) describes a collision with the ego's driver-sided door. Lastly, Case (V) considers an impact towards the back of the ego. For each test case, we keep the uncertainty, i.e., $\sigma_x, \sigma_y, \sigma_\theta, \sigma_v$, constant. The motion of each vehicle is modeled by constant velocity model, and each vehicle's shape is approximated by three circles. 
The three-circle approximation of each actor's shape yields nine severity weighting factors $w_{j,l}$, that we collect in a matrix $W$. We assign the weights such that front-to-side collisions are weighted the strongest, followed by frontal, and, lastly, rear collisions. All simulation parameters are provided in Table \ref{tab:parameters}.

\subsection{Computational Performance}
As the risk function is based on a method for collision probability estimation, we compare the computational performance of the risk function to the algorithm for collision probability estimation of~\cite{Tolksdorf_POC_2024}. 
To estimate the risk we use the implementation of~\cite{Tolksdorf_POC_2024}, however, we substitute Algorithm 3 "Sort Intervals 'SortInts'" in~\cite{Tolksdorf_POC_2024} with our Algorithm \autoref{alg:weigthed_severity}, to calculate risk instead. Note that both resulting algorithms consist of an initialization step and an estimation step. The initialization step only has to be performed once for a given geometry and number of circles of both vehicles, whereas the estimation step is typically carried out at high frequency at runtime (e.g., in a motion planning context). The initialization step calculates the disjoint intersection angle intervals $\breve{\mathbb{I}}^{\theta}(\phi, \rho)$ and, for the risk estimation algorithm, the expected conditional severity per interval $\mathbb{E}[s_m(\cmbtheta, v_e, v_o) \mid v_o]$. Once initialized, the estimation step is only left with computing (\ref{eq:analytic_warpped_gauss}) and (\ref{eq:ego_risk_short}). 
We calculate the average initialization time over 100 initializations and the average estimation time over $10^4$ estimations. We measure both times for a two, three, and four circle vehicle approximation. Furthermore, we randomly vary both vehicle configurations, velocities, and uncertainties between each function call, so as not to calculate the same risk and collision probability for each measurement. 
As a computational platform we use a notebook using an Intel Core i7-12800HX processor clocked at up to $4.80\,\text{GHz}$ with $32\,\text{GB}$ of memory.

\subsection{Results}
\begin{figure}
    \centering
    \begin{tikzpicture}
        \begin{axis}[xlabel={Time [s]}, ylabel={Risk}, no markers, every axis plot post/.append style={line width=1.5pt}, xmin=-0.5, xmax=2, ymin=0, ymax=400000, width=1.01\linewidth, grid=major]
            \draw [solid, line width=0.9pt,black] (axis cs: 1, 0) -- (axis cs: 1, 59492);
            \draw [solid, line width=0.9pt,black] (axis cs: 0.5, 0) -- (axis cs: 0.5, 55425);
            \draw [solid, line width=0.9pt,black] (axis cs: 0.5, 0) -- (axis cs: 0.5, 400000);
            \draw [solid, line width=0.9pt,black] (axis cs: 1, 0) -- (axis cs: 1, 400000);

            \addplot+[myblue] table [x expr=0.001*\thisrow{time_ms}-1.25,y=ego_risk,col sep=comma] {NumericalCaseData/res_default_head_on_edit.json.csv};
            \addplot+[dashed, myred] table [x expr=0.001*\thisrow{time_ms}-1.5,y=ego_risk,col sep=comma] {NumericalCaseData/res_default_rear_end_edit.json.csv};
            \addlegendentry{Head-on (I)};
            \addlegendentry{Rear-end (II)};
            \filldraw (axis cs: 1,9492) circle (1.3pt);
            \filldraw (axis cs: 0.5, 5425) circle (1.3pt);
        \end{axis}
        \node at (5.5+0.4,2.1) {\includegraphics[scale=0.65]{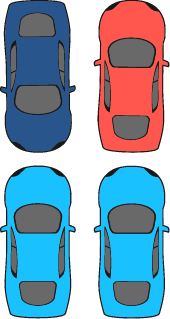}};
        \node[anchor=center] at (4.98+0.4, 4.2) {(I)};
        \node[anchor=center] at (6.02+0.4, 4.2) {(II)};
    \end{tikzpicture}
    \caption{Cases (I): the object (dark blue vehicle) is colliding with the ego (light blue vehicle) head-on. Case (II): The ego collides into the object's (red vehicle) rear.
    }
    \label{fig:NumCaseHeadOnRearEnd}
\end{figure}
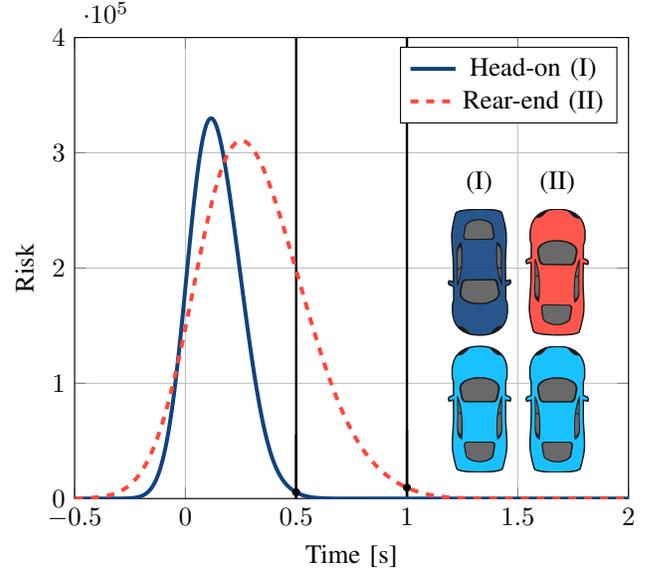

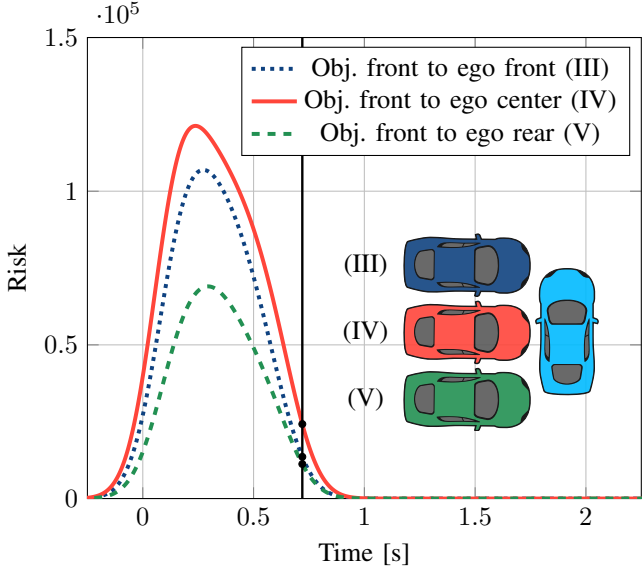
\begin{figure}
    \centering
    \begin{tikzpicture}
        \begin{axis}[xlabel={Time [s]}, ylabel={Risk}, no markers, every axis plot post/.append style={line width=1.5pt}, xmin=-0.25, xmax=2.25, ymin=0, ymax=150000, width=1.01\linewidth, grid=major,
        legend style={
        at={(0.98,0.98)},       
        anchor=north east 
    }]

            \addplot+[dotted, myblue] table [x expr=0.001*\thisrow{time_ms}-0.72,y=ego_risk,col sep=comma] {NumericalCaseData/res_default_side_front.json.csv};
            \addplot+[myred] table [x expr=0.001*\thisrow{time_ms}-0.72,y=ego_risk,col sep=comma] {NumericalCaseData/res_default_side_mid.json.csv};
            \addplot+[mygreen, dashed, mygreen] table [x expr=0.001*\thisrow{time_ms}-0.72,y=ego_risk,col sep=comma] {NumericalCaseData/res_default_side_rear.json.csv};
            \addlegendentry{Obj. front to ego front (III)};
            \addlegendentry{Obj. front to ego center (IV)};
            \addlegendentry{Obj. front to ego rear (V)};

            \draw [solid, line width=0.9pt,black] (axis cs: 0.72, 0) -- (axis cs: 0.72, 150000);
            \filldraw [black] (axis cs: 0.72,24226.6) circle (1.3pt);
            \filldraw [black] (axis cs: 0.72,13575.4) circle (1.3pt);
            \filldraw [black] (axis cs: 0.72,11234.1) circle (1.3pt);
        \end{axis}

        \node at (5.5,2.21) {\includegraphics[scale=0.65]{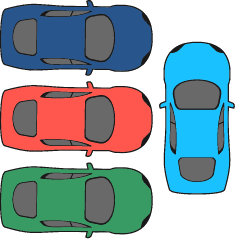}};
        \node[anchor=center] at (3.7, 3.10) {(III)};
        \node[anchor=center] at (3.7, 2.21) {(IV)};
        \node[anchor=center] at (3.7, 1.32) {(V)};

    \end{tikzpicture}
    \caption{Side collision Cases (III) - (V), the ego is displayed in light blue and the object's color is different in each case.}
    \label{fig:NumCaseSides}
\end{figure}

\paragraph{Risk estimation for varying collision constellations} The results for all test cases are given in~\autoref{fig:NumCaseHeadOnRearEnd} and \autoref{fig:NumCaseSides}. Here, we let the vehicles traverse through the respective Cases (I) - (V) and estimate the risk, after a single initialization for each test case, at a frequency of \qty{1}{ms}. The graphs are aligned such that at \qty{0}{s} both vehicles touch for the first time for the mean motion of the object, i.e., the collision begins. Regarding  Cases (I) and (II), we report maximum risk values of $3.3 \cdot 10^5$ in the head-on case and $3.1 \cdot 10^5$ in the rear-end case. Note that both vehicles are traveling through each other, and, hence, at collision impact, i.e., at \qty{0}{s}, the rear-end case reports a risk of $1.5 \cdot 10^5$ and the head-on case is estimating a considerably higher risk at $2.0 \cdot 10^5$. \\
The vertical markings over the $x$-axis in Figures~\ref{fig:NumCaseHeadOnRearEnd} and~\ref{fig:NumCaseSides} give the respective collision endings for the mean object motion, i.e., at that time, both vehicles stop colliding. 
The head-on collision reports the highest maximum risk shortly after the start of the collision, as (\ref{eq:analytic_severity}) measures the head-on case with the absolute kinematic energy, and a collision between a front and a center circle is weighted the strongest (which starts to occur shortly after the collision begins). The graph of the head-on cases rises later as the rear-end case due to the higher relative velocity, i.e., the distance between both vehicles reduces in less time.
Note that in the rear-end case, a head-on collision is also occurring when the ego travels through the object (hence, both front circles intersect at a later point); however, here both vehicles are \textit{also} colliding with their other, lower-weighted sections, where the total severity value is then averaged. This combination explains the delayed, slightly lower peak in the rear-end case. Whether such behavior is desirable depends on the application of the risk estimation, e.g., one may restrict risk estimation to some brief interval until the start of a collision.\\ 
We report a similar effect for Cases (III)-(V), in which the different criticalities of potential collisions become visible. Associated with the highest risk is Case (IV) at $1.21\cdot 10^5$, followed by Case (III) with $1.07\cdot 10^5$ and, finally, Case (V) at $0.69\cdot 10^5$. The reduction of about $12\,\%$ from side-to-front collisions and another $64\,\%$ from front-to-rear impacts showcases the design of the severity matrix $W$, that weighs front-to-side collisions as the most severe.\\
\paragraph{Computational performance} \autoref{tab:executionTimesRiskPOC} reports the average initialization and estimation times for risk and collision probability calculations, as well as their difference. For the initialization time, we find an increase by a factor of about $4.4$ for the circle added to the ego and the object. Concerning risk and collision probability estimation times, a speed-up of $60\,\text{\%}$ is found for reducing the number of circles used for vehicle approximation per vehicle from three to two circles. Meanwhile, the differences between risk and collision probability calculation times are negligibly low at maximum $8.13\,\mu\text{s}$. Note that for many applications, e.g., motion planning, the initialization times are inconsequential, as collision probability or risk estimation algorithms are initialized before run-time. Furthermore, it is possible to prepare multiple risk functions, which are then only called from memory instead of being initialized whenever needed. Given the fact that both risk and collision probability estimation are calculated in little time, both algorithms can be suitable for use in real-time motion planning.

\begin{table}[h]
    \centering
    \caption{Calculation times for initialization, risk, and probability of collision (POC).}
    \begin{tabular}{c|ccc|c}
        \# Circles & Init. [ms] & Risk [ms] & POC [ms] & $\text{POC}-\text{Risk}$ [ms]\\ \hline
        2 & 1641.65 & 0.25761 & 0.25177 & 0.00584\\
        3 & 6472.96 & 0.43018 & 0.42205 & 0.00813\\
    \end{tabular}
    \label{tab:executionTimesRiskPOC}
\end{table}

\section{Conclusions and Future Work}\label{sec:conclusion}
In this paper, we propose a risk estimation methodology for automated vehicles, in which risk reflects both the probability and severity of a collision event. The risk estimation approach is shown have good computational performance and is hence suitable to be deployed in, e.g., motion planning algorithms or for general safety assessment. The proposed method allows us to assign different collision severities for different collision constellations. To achieve this, we extend the method for collision probability estimation of \cite{Tolksdorf_POC_2024} to risk estimation, inheriting us computational efficiency and estimation accuracy as we utilize the same algorithmic structure. Furthermore, we provide an exemplary derivation for risk estimation for the scenario of one vehicle facing another vehicle with Gaussian uncertainties in the position, heading angle, and velocity, along with a kinematic energy model that considers the collision severity differently between different collision constellations. Given the severity model and Gaussian uncertainties, we can reduce the computational effort of risk estimation significantly by reducing the four-dimensional integral to a two-dimensional integral, allowing us to leverage established efficient numerical integration methods.
We demonstrate in a representative example that our risk estimation approach indeed allows us to assign different collision severities for different collision constellations. In addition, we show that our algorithm is computationally efficiently enough for real-time motion planning applications. Lastly, we provide the programming code for the risk function as open source.\\
For future work, we envision to investigate other models for the uncertainties, which we assumed to be Gaussian for exemplary purposes and due to their frequent occurrence. Similarly, other and higher-fidelity severity models may be employed to represent collision severity more accurately. Here, we note that especially the resulting combination of probability density functions and severity functions is interesting, as the computational efficiency depends on the functional forms of both, i.e., whether arising integrals can be analytically solved. As our results demonstrate the efficacy with respect to computational efficiency and severity estimation under different collision constellations, we clearly recommend the use for motion planning applications. 

\begin{table}
    \centering
    \caption{Parameters for the numerical case study.}
    \begin{tabular}{c|ccccc}
                             & Case (I) & Case (II) & Case (III) & Case (IV) & Case (V)\\\hline
        $x_{e,k=0}$                &  -15 & -15 &   0 & 0 & 0\\
        $y_{e,k=0}$                &    0 &   0 &   0 & 0 & 0\\
        $\theta_{e,k=0}$           &    0 &   0 & $-\pi/2$ & $-\pi/2$ & $-\pi/2$\\
        $v_e$                &   15 &  15 & 0 & 0 & 0\\\hline
        $\mu_{x,k=0}$          &   15 &   5 & -15 & -15 & -15\\
        $\mu_{y,k=0}$          &    0 &   0 & -3 & 0 & 3\\
        $\mu_{\theta,k=0}$     &$\pi$ &   0 & 0 & 0 & 0\\
        $\mu_{v}$          &    5 &   5 & 13.89 & 13.89 & 13.89\\\hline
        $\sigma_{x}$       &  1.5 & 1.5 & 1.5 & 1.5 & 1.5\\
        $\sigma_{y}$       &  1.5 & 1.5 & 1.5 & 1.5 & 1.5\\
        $\sigma_{\theta}$  &  1.5 & 1.5 & 1.5 & 1.5 & 1.5\\
        $\sigma_{v}$       &  1.5 & 1.5 & 1.5 & 1.5 & 1.5\\\hline
        $\underline{v}_{o}$  &    0 &   0 & 10 & 10 & 10\\
        $\overline{v}_{o}$   &   10 &  10 & 15 & 15 & 15\\
    \end{tabular}
    \label{tab:Case_Study_Parameters}
\end{table}

\begin{table}
    \centering
    \caption{Approximation configuration for the numerical case study.}
    \begin{tabular}{r| cc|cc}
       Ego \& object masses & $m_e $ & 1000 & $m_o$ & 1000\\
       Ego \& object lengths& $l_e$ & 5 & $l_o$ & 5\\
       Ego \& object widths& $w_e$ & 2.2 & $w_e$ &2.2\\
       Ego \& object no. of circles& $N_e$ & 3 & $N_o$ & 3\\\hline
       Severity weighting matrix & \multicolumn{1}{c}{$W$} & \multicolumn{3}{l}{$\vphantom{\begin{pmatrix} a \\ a \\ a \\ a\end{pmatrix}}\begin{pmatrix} 5 & 20 & 1 \\
         20 & 1 & 1 \\
         1 & 1 & 1\end{pmatrix}$}
    \end{tabular}
    \label{tab:parameters}
\end{table}

\newpage
\bibliography{lib}
\bibliographystyle{ieeetr}

\end{document}